%% file: arxiv.tex
\documentclass{article} 
\usepackage{arxiv}

\usepackage{graphicx}
\usepackage{booktabs}
\usepackage[accsupp]{axessibility}  
\usepackage[pagebackref,breaklinks]{hyperref}
\usepackage{orcidlink}
\usepackage{colortbl}
\definecolor{light-light-gray}{gray}{0.92} 
\usepackage{wrapfig}
\usepackage{algorithm}
\usepackage{algpseudocode}
\usepackage[normalem]{ulem}
\usepackage{array}

\input{math_commands.tex}

\usepackage{graphicx}
\usepackage{hyperref}
\usepackage{url}
\usepackage{booktabs}
\usepackage{multirow} 
\usepackage{booktabs}
\usepackage{makecell}
\usepackage[framemethod=default]{mdframed}
\usepackage[strict]{changepage}
\usepackage{enumitem}

\newcolumntype{R}[1]{>{\raggedleft\arraybackslash}p{#1}}

\newmdenv[
  linewidth=0pt,
  linecolor=black,
  innerleftmargin=5pt,
  innerrightmargin=5pt,
  skipabove=5pt,
  skipbelow=5pt
]{promptbox}

\newmdenv[
  linewidth=1pt,
  linecolor=black,
  topline=true,
  bottomline=true,
  leftline=true,
  rightline=true,
  innerleftmargin=10pt,
  innerrightmargin=10pt,
  innertopmargin=10pt,
  innerbottommargin=10pt,
  skipabove=1pt,
  skipbelow=1pt
]{examplebox}

\title{\emph{CellFluxRL}: Biologically-Constrained Virtual Cell Modeling via Reinforcement Learning}

\author{ {\hspace{0.1mm}Dongxia Wu}$^{*}$\\
    \texttt{Stanford University}\\
    \texttt{Stanford, CA}\\
	\texttt{dowu@stanford.edu} \\
    \And{\hspace{0.1mm}Shiye Su}$^{*}$\\
    \texttt{Stanford University}\\
    \texttt{Stanford, CA}\\
	\texttt{shiye@stanford.edu} \\
    \And{\hspace{0.1mm}Yuhui Zhang}$^{*}$\\
    \texttt{Stanford University}\\
    \texttt{Stanford, CA}\\
    \texttt{yuhuiz@stanford.edu}\\
    \And{\hspace{0.1mm}Elaine Sui}\\
    \texttt{Stanford University}\\
    \texttt{Stanford, CA}\\
	\texttt{esui@stanford.edu} \\
    \And{\hspace{0.1mm}Emma Lundberg}\\
    \texttt{Stanford University}\\
    \texttt{Stanford, CA}\\
	\texttt{emmalu@stanford.edu} \\
    \And{\hspace{0.1mm}Emily B. Fox$^{\dagger}$}\\
    \texttt{Stanford University}\\
    \texttt{Stanford, CA}\\
	\texttt{ebfox@stanford.edu} \\
    \And{\hspace{0.1mm}Serena Yeung-Levy}$^{\dagger}$\\
    \texttt{Stanford University}\\
    \texttt{Stanford, CA}\\
    \texttt{syyeung@stanford.edu} \\
}

\date{}

\newcommand{\blfootnote}[1]{\begingroup%
\renewcommand\thefootnote{}\footnotetext{#1}%
\addtocounter{footnote}{-1}%
\endgroup}

\begin{document}

\maketitle

\blfootnote{$^*$ Equal contribution. $^\dagger$ Equal advising.}

\input{secs/0_abs}

\input{secs/1_intro}

\input{secs/5_related}

\input{secs/2_problem}

\input{secs/3_method}
\input{secs/4_exp_arxiv}

\section*{Acknowledgement}
This work was supported in part by ONR Grant N00014-22-1-2110, NSF Grant 2205084, and the Stanford Institute for Human-Centered Artificial Intelligence (HAI). EBF is a Biohub, San Francisco, Investigator. S.Y. is a Chan Zuckerberg Biohub — San Francisco Investigator. E.L. was supported by the Stanford Institute for Human-Centered AI.

\bibliography{main}
\bibliographystyle{plain}

\newpage
\appendix
\input{secs/X_appendix_arxiv}

\end{document}

%% file: math_commands.tex
\usepackage{amsmath,amsfonts,bm}

\def\eqref#1{equation~\ref{#1}}

\def\1{\bm{1}}

\DeclareMathAlphabet{\mathsfit}{\encodingdefault}{\sfdefault}{m}{sl}
\SetMathAlphabet{\mathsfit}{bold}{\encodingdefault}{\sfdefault}{bx}{n}



%% file: secs/0_abs.tex
\begin{abstract}

Building virtual cells with generative models to simulate cellular behavior in silico is emerging as a promising paradigm for accelerating drug discovery. However, prior image-based generative approaches can produce implausible cell images that violate basic physical and biological constraints. 
To address this, we propose to post-train virtual cell models with reinforcement learning (RL), leveraging biologically meaningful evaluators as reward functions.
We design seven rewards spanning three categories—biological function, structural validity, and morphological correctness—and optimize the state-of-the-art \textit{CellFlux} model to yield \textit{CellFluxRL}.
\emph{CellFluxRL} consistently improves over \emph{CellFlux} across all rewards, with further performance boosts from test-time scaling. 
Overall, our results present a virtual cell modeling framework that enforces physically-based constraints through RL, advancing beyond “visually realistic” generations towards “biologically meaningful” ones.

\end{abstract}

%% file: secs/1_intro.tex
\section{Introduction}
\label{sec:intro}

Wet-lab experiments have long been the primary bottleneck in drug discovery. To conduct a single experiment, researchers must not only purchase expensive reagents, consumables, and equipment, but also wait weeks or even months to synthesize drugs and culture cells. Consequently, biologists have long envisioned building virtual cells~\cite{slepchenko2003quantitative,johnson2023building,bunne2024build} to accelerate this process by simulating cell behavior \textit{in silico}. With recent advances in generative modeling, such as diffusion models~\cite{sohl2015deep,song2019generative,ho2020denoising} and flow matching~\cite{lipmanflow,lipman2024flow,liu2024flowing,liu2022flow}, alongside high-throughput image-based screening technologies~\cite{chandrasekaran2023jump,chandrasekaran2024three,fay2023rxrx3} that produce terabytes of data, there is growing interest in developing image-based virtual cells to predict morphological responses to perturbations.

\begin{wrapfigure}{r}{0.2\textwidth}
    \centering
    \vspace{-2.2em}
    \includegraphics[width=0.2\textwidth]{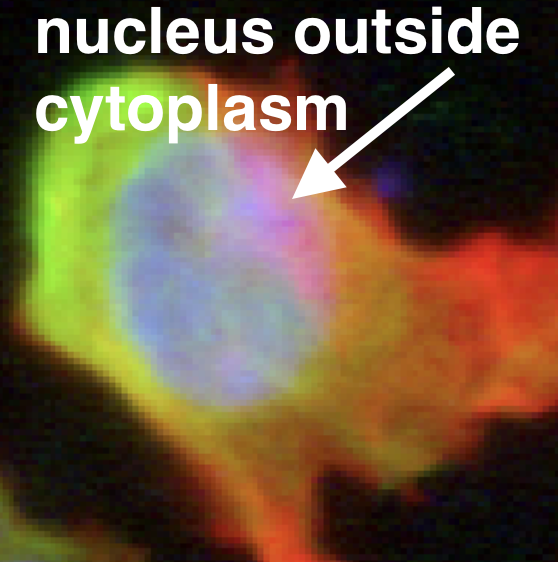}
    \vspace{-1.9em}
    \caption{Failure of cell generation.}
    \vspace{-3.5em}
    \label{fig:motivation}
\end{wrapfigure}

Despite its success, we observe that these image-based virtual cell models can produce images that look realistic yet are biologically implausible. For instance, using the state-of-the-art image-based virtual cell model, \emph{CellFlux}~\cite{zhang2025cellflux}, we observe anomalies such as the cell nucleus being generated outside of the cytoplasm (Figure~\ref{fig:motivation}). Such violations greatly limit real-world use of virtual cell models, reducing their practical value.



\begin{figure}[!tb]
    \centering
    \includegraphics[width=0.95\linewidth]{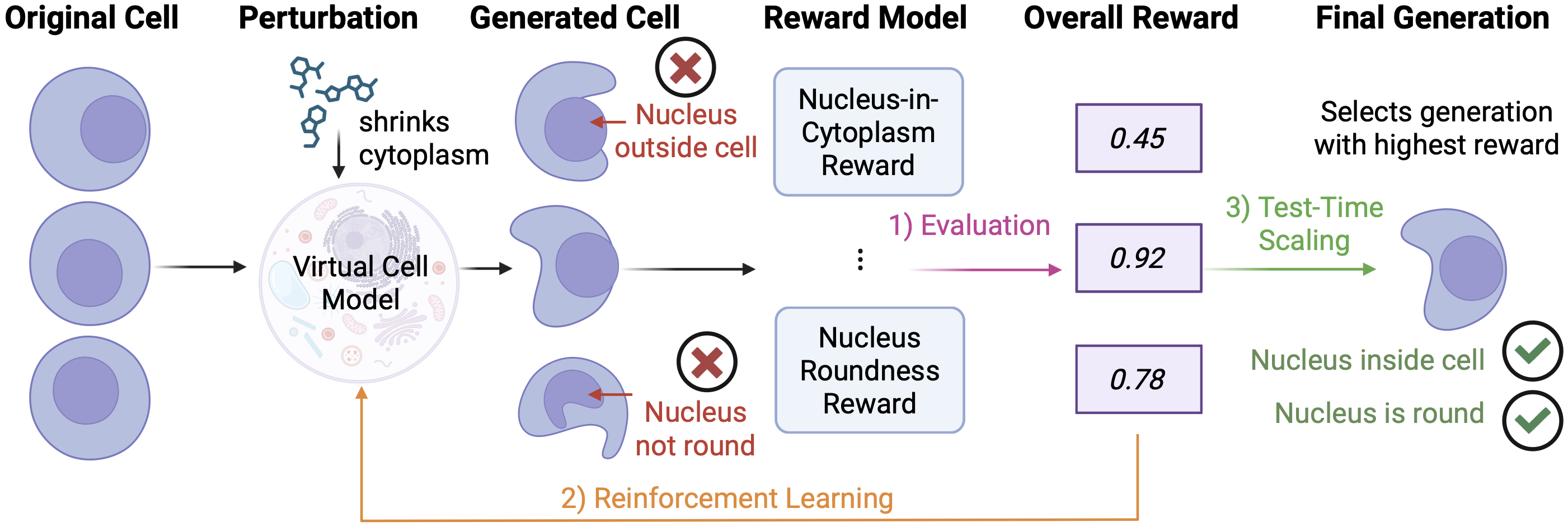}
    \caption{\textbf{Motivation.} Current generative models for simulating cellular perturbations can fail to produce physically plausible cell images. For example, nuclei may appear outside the cell membrane. We design a suite of biologically meaningful verifiers in three roles: (1) as \emph{evaluators} to assess the biological correctness of generated images, (2) as \emph{reward signals} to improve generation via reinforcement learning, and (3) as \emph{verification modules} to enhance sample quality through test-time scaling.}
    \label{fig:pull}
\end{figure}

We hypothesize that these failures arise from a mismatch between the pixel-level flow-matching objective and global physical constraints. Flow-based generative models~\cite{lipman2024flow} learn velocity fields that transport samples from one distribution to another. While this objective can encourage visual fidelity, it does not explicitly enforce higher-level constraints on cell structure and properties. This may yield samples that appear plausible locally but are incorrect globally.

To address this issue, we propose \emph{CellFluxRL}, applying reinforcement learning (RL)-based post-training with biologically-meaningful evaluators as reward functions. Concretely, many constraints of interest admit existing or readily obtained evaluators that can efficiently assign scores to generated images (e.g., whether a sample satisfies structural or functional criteria). Although these evaluators are typically non-differentiable and thus cannot be optimized via standard backpropagation, they can be used as reward signals for RL~\cite{liu2025flow,zheng2025diffusionnft}. At a high level, our RL procedure alternates between: (1) \textbf{sampling}, where we generate multiple images and evaluate them using biologically-meaningful evaluators; and (2) \textbf{optimization}, where we increase the likelihood of high-reward samples and decrease the likelihood of low-reward samples. This directly penalizes physically invalid generations while reinforcing physically consistent ones. Although our work focuses on improving \emph{CellFlux}, a flow-based model, the same training ideas can be applied to other generative modeling frameworks.

To demonstrate the potential of this approach, we post-train \textit{CellFlux} to yield \emph{CellFluxRL}, optimizing seven reward models that assess correctness across three key categories: \textbf{biological function}, such as mode of action; \textbf{structural properties}, including nuclear roundness and containment of the nucleus within the cytoplasm; and \textbf{morphological properties}, like the size and number of nuclei and the size of the cytoplasm. Our overall reward is a weighted linear combination of component rewards. Evaluating the learned model across all reward components, \emph{CellFluxRL} achieves higher scores than the base model. These results indicate that RL provides an effective mechanism for aligning image-based generative models with biologically-meaningful constraints. We additionally present qualitative case studies and ablations to analyze how RL hyperparameters and reward design influence performance.

Moreover, these reward models enable a simple form of test-time scaling. Following the selection strategy of~\cite{ma2025inference}, we sample $N$ candidate images for a given condition and select the one with the highest reward. We observe monotonic improvements as $N$ increases, indicating a clear test-time scaling trend, and achieving an additional gain over single-sample generation.

Our work also shines a light on limitations of existing evaluation metrics for image-based generative models of cellular morphology. To date, focus has been on image generation quality scores, like FID and KID, rather than biological validity. Our considered rewards not only enable RL-based training, but also new, biologically meaningful evaluation criteria.

In summary, our work identifies a key limitation of existing image-based virtual cell modeling: the lack of explicit enforcement of physical correctness under pixel-level training objectives. We address this limitation by incorporating biologically-meaningful evaluators through RL, which substantially improves physical plausibility of generated images. We further show that these gains can be amplified via test-time scaling. Finally, our rewards serve as new metrics for the community to benchmark against. Overall, our results move virtual cell image generation from “looking good” to being \emph{biologically consistent}, supporting the downstream goal of drug discovery and personalized medicine with reduced reliance on costly wet-lab experiments.

%% file: secs/5_related.tex
\section{Related Work}
\label{sec:related_work}

\paragraph{Virtual cell modeling.} A \emph{virtual cell} model aims to simulate cellular responses to perturbations \emph{in silico}. This can be formulated as a generative modeling problem $p(x_1|x_0, c)$, where $x_0$ and $x_1$ denote cell states before and after perturbation $c$. Early virtual cell models relied on mathematical and physics-based frameworks, such as systems of ordinary differential equations (ODEs)~\cite{you2004toward,walkerthevirtualcell,SLEPCHENKO20101}. However, their expressiveness is inherently limited by the strict assumptions and simplified dynamics that these formulations require. With recent advances in generative modeling and image-based high-throughput screening, a growing body of work has explored using VAEs, GANs, diffusion models, and flow matching to simulate cellular state transitions~\cite{bereketmodelling,10.1093/bioinformatics/btz158,lamiable2023revealing,palma2023predicting,bourou2024phendiff,hung2024lumic,zhang2025cellflux}. A notable paradigm is \emph{CellFlux}~\cite{zhang2025cellflux}, which reformulates the problem as a distribution-to-distribution transformation and leverages flow matching to mitigate batch effects, achieving state-of-the-art performance. Nevertheless, we observe that while these models generate visually plausible cells, the generations can violate fundamental biological constraints. To address this challenge, we introduce RL post-training guided by biologically meaningful rewards, encouraging virtual cell models to generate more physically and biologically plausible cell images.

\paragraph{Physics-aware generative models.} The growing interest in utilizing generative models as ``world simulators'' or ``world models'' necessitates strict adherence to physical accuracy~\cite{chen2021transdreamer,MATSUO2022267,pmlr-v205-wu23c}. However, recent studies indicate that state-of-the-art image and video generative models still struggle to maintain physical correctness, a limitation that cannot be resolved simply by scaling up data and model parameters~\cite{meng2024towards, kang2025far}. To tackle this challenge, the community has proposed various approaches: modifying architectures to inject physical constraints~\cite{Huang_2023_CVPR,Yang_2024_CVPR}, altering the inference process by using LLMs for generation planning~\cite{Lin2023VideoDirectorGPT,NEURIPS2024_edbeca78}, or modifying training objectives to incorporate physical preferences~\cite{xu2024visionrewardfinegrainedmultidimensionalhuman,liu2025improving,qian2025rdporealdatapreference,cai2026phygdpophysicsawaregroupwisedirect}. Our approach aligns with objective-based methods; in virtual cell modeling, we can readily define numerous biologically meaningful constraints. Optimizing these constraints requires minimal architectural modifications while yielding robust results. Furthermore, these rewards facilitate test-time scaling to continuously enhance generation quality.

\paragraph{Reinforcement learning for generative modeling.} While generative models based on diffusion or flow matching achieve high visual fidelity, aligning them with strict physical constraints remains an open challenge. RL has emerged as a powerful solution~\cite{black2023training, fan2023dpok, liu2025flow, 
xue2025dancegrpo,li2025mixgrpo}, yet its development is still in its nascent stages. The primary hurdle lies in evaluating the exact likelihood of samples, necessary for adjusting generation probabilities based on rewards, which is generally intractable in diffusion and flow matching frameworks. Prior works, such as FlowGRPO~\cite{liu2025flow} and DanceGRPO~\cite{xue2025dancegrpo}, formulate this as a Markov Decision Process (MDP) to derive probabilities, whereas recent algorithms like DiffusionNFT~\cite{zheng2025diffusionnft} introduce forward processes to estimate these likelihoods. Our work builds upon the DiffusionNFT-style forward probability estimation process, which offers rapid estimation and state-of-the-art results. Crucially, it accommodates our distribution-to-distribution flow matching (from original to perturbed states), whereas MDP-based approaches assume a standard Gaussian prior.

%% file: secs/2_problem.tex
\section{Cellular Perturbation Modeling}

Our approach takes as input a pretrained generative model for cellular morphology prediction and 
post-trains it using RL to improve physical correctness. 
In this section, we describe the 
perturbation modeling problem and the properties required for the base model.



\paragraph{Objective.}
Let $\mathcal{X} \subseteq \mathbb{R}^{H \times W \times C}$ denote the space of multi-channel fluorescence 
microscopy images, where each channel highlights a distinct cellular structure (e.g., nucleus, 
cytoskeleton, mitochondria). Let $\mathcal{C}$ denote the space of perturbations, encompassing chemical compounds (drugs). Given an 
unperturbed cell image $x_0$ and a perturbation $c \in \mathcal{C}$, the goal is to learn a 
generative model that samples from the conditional distribution $p(x_1 \mid x_0, c)$, where $x_1$ 
represents the cell's morphology after treatment. Such a model enables \textit{in silico} simulation 
of cellular responses that would otherwise require costly wet-lab experiments.

\paragraph{Data and batch effects.}
Cell morphology data are collected via high-content microscopy screening, in which multi-well plates are 
prepared with both \textit{control} wells (untreated) and \textit{perturbed} wells (treated with a chemical compound). Experiments are conducted across multiple batches, each introducing 
systematic technical variations---differences in staining intensity, illumination, or imaging conditions---that are unrelated to the perturbation itself. Because imaging is destructive (cells are fixed and stained), 
paired before-and-after observations of the same cell are unavailable. Instead, we observe unpaired sets of 
control images $\{x_0\}$ and perturbed images $\{x_1\}$ within each batch.

\paragraph{Requirements for the base model.}
These data characteristics impose two key requirements on the base generative model. First, because 
observations are unpaired, the model must learn a \textit{distributional} transformation from control 
to perturbed cells rather than a pointwise mapping. Second, because batch effects confound perturbation 
signals, the model should condition on same-batch control images as its source distribution, so that it 
learns only the perturbation-induced changes rather than technical artifacts. 
While our approach is compatible 
with any such base model, we instantiate it as \emph{CellFlux}~\cite{zhang2025cellflux}.
\emph{CellFlux}, based on flow matching, offers a natural framework satisfying the above requirements: 
it learns a velocity field $v_\theta\colon \mathcal{X} \times [0,1] \times \mathcal{C} \to \mathcal{X}$ 
that continuously transports the control distribution $p_0$ to the perturbed distribution $p_1(\cdot \mid c)$ 
within each batch. Training proceeds by sampling pairs $(x_0, x_1)$ from source and target distributions, 
constructing linear interpolations $x_t = (1-t)\,x_0 + t\,x_1$ for $t \sim \mathcal{U}[0,1]$, and minimizing:
\begin{equation}
\label{eq:flow_matching}
\mathcal{L}_{\mathrm{FM}}(\theta) = \mathbb{E}_{x_0 \sim p_0,\, x_1 \sim p_1,\, t \sim \mathcal{U}[0,1]} \bigl\| v_\theta(x_t, t, c) - (x_1 - x_0) \bigr\|_2^2
\end{equation}
At inference, a control image $x_0$ is transformed into a predicted perturbed image $\hat{x}_1$ by solving 
the ODE $\mathrm{d}x_t = v_\theta(x_t, t, c)\,\mathrm{d}t$ from $t{=}0$ to $t{=}1$. We denote the resulting 
pretrained model as $v_\theta$, which serves as the input to our method.

%% file: secs/3_method.tex
\section{Method}
\label{sec:method}

In this section, we present our approach to improving the physical and biological fidelity of virtual cell models. We first introduce the formal objective for reward-driven generation (\S\ref{sec:reward_driven_generation}). Next, we detail the suite of reward models/evaluators (\S\ref{sec:reward_models}) designed to capture biological and structural properties. We then describe the RL algorithm (\S\ref{sec:rl_algorithm}) used to post-train the base model via contrastive updates. Finally, we discuss test-time scaling (\S\ref{sec:test_time_scaling}) to further enhance generation quality at inference time.

\subsection{Reward-Driven Generation}
\label{sec:reward_driven_generation}

While the flow matching objective trains the base model to produce images that are distributionally close to real perturbed cells, it operates at the pixel-level and does not explicitly encourage higher-level physical or biological correctness. In practice, this can lead to generated images that appear visually plausible but violate known cellular properties---for example, producing nuclei of incorrect size, cells with implausible morphology, or images that fail to reflect the expected mode of action (MoA) of a compound (Figure~\ref{fig:pull}). Such violations limit the utility of virtual cell models for downstream applications like drug screening, where physical correctness is essential.

We observe that many properties of interest can be assessed by \textit{evaluator functions} that score generated images along specific axes of physical or biological validity. Formally, let $\{r_k\}_{k=1}^K$ denote a set of reward functions, where each $r_k\colon \mathcal{X} \times \mathcal{C} \to \mathbb{R}$ measures how well a generated image $\hat{x}_1$ satisfies a particular property given perturbation $c$. These rewards capture complementary aspects of cell structure and function, such as whether the nuclear size is consistent with the expected effect of the perturbation, or whether the overall morphological profile matches the correct MoA. We describe the specific reward functions used in this work in \S\ref{sec:reward_models}.

Given the pretrained base model $v_\theta$, our objective is to obtain a post-trained model $v_{\theta'}$ that generates images with higher physical fidelity as measured by $\{r_k\}$, while remaining close to the pretrained model. Concretely, we seek to solve:
\begin{equation}
\label{eq:rl_objective}
\max_{\theta'} \; \mathbb{E}_{x_0 \sim p_0,\, c \sim p(c)} \left[ \sum_{k=1}^K w_k r_k(\hat{x}_1, c) \right] \quad \text{s.t.} \quad D_\text{KL}(v_{\theta'} \| v_\theta) \leq \epsilon
\end{equation}
where $\hat{x}_1$ is the image generated by $v_{\theta'}$ from input $(x_0, c)$, $w_k$ is the weight of each reward $r_k(.)$, and $D_\text{KL}(\cdot \| \cdot)$ is the KL divergence between the post-trained and pretrained models. This regularization is important both to preserve the visual quality and diversity learned during pretraining, and to mitigate \textit{reward hacking}---generating images that achieve high scores through degenerate solutions rather than genuine physical correctness.

Critically, optimizing these rewards is complementary to, and not in conflict with, the original distributional objective. The reward functions encode properties that the \textit{ground truth} perturbed distribution satisfies. For instance, real cells treated with a microtubule destabilizer exhibit smaller, fragmented nuclei, and real cells have mode-of-action-consistent morphological profiles. By directly optimizing for these properties, RL can improve the match between generated and real distributions along biologically meaningful axes that the pixel-level flow matching loss may underweight. In this sense, the rewards provide a form of \textit{targeted distributional alignment}, focusing model capacity on the aspects of the distribution that matter most for physical correctness.

\subsection{Reward Models}
\label{sec:reward_models}

We design a suite of reward functions that capture complementary aspects of biological validity in generated cell images. These rewards fall into three categories: \textit{biological function} rewards that assess whether generated images reflect the expected biological effects of a perturbation, \textit{structural} rewards that enforce known physical relationships between cellular components, and \textit{morphological} rewards that encourage generated images to match the size and density statistics of real cells. Together, these rewards provide multi-scale supervision---from high-level biological semantics down to low-level geometric properties---that the pixel-level flow matching objective does not explicitly optimize for.

\subsubsection{Biological Function.}
\label{sec:reward_biological}

Biological function rewards evaluate whether generated images faithfully reflect the biological effects of 
the applied perturbation.

\paragraph{Mode of action (MoA).}
A drug's MoA describes the cellular process it targets, such as microtubule destabilization, DNA replication inhibition, or actin disruption. Because different MoAs produce distinct morphological signatures, the ability of a generated image to be correctly classified by MoA serves as a strong indicator of biological fidelity. We leverage an MoA classifier pretrained on real perturbed images~\cite{zhang2025cellflux} and define the reward as the predicted probability of the 
ground truth MoA class:
\begin{equation}
\label{eq:reward_moa}
r_{\mathrm{MoA}}(\hat{x}_1, c) = p_{\mathrm{cls}}(y_c \mid \hat{x}_1),
\end{equation}
where $y_c$ is the ground truth MoA label associated with perturbation $c$ and $p_{\mathrm{cls}}$ is the pretrained MoA classifier. Higher values indicate that the generated image exhibits the morphological profile expected for the given perturbation.


\subsubsection{Structural Constraints.}
\label{sec:reward_structural}

Structural constraint rewards enforce known physical relationships between cellular components.

\paragraph{Nucleus-in-cytoplasm.}
In correctly imaged cells, nuclei are fully enclosed within the cytoplasm. Generated images that violate this containment
are physically implausible. We segment nuclei and cytoplasm from the generated images using Cellpose~\cite{pachitariu2022cellpose} and define the reward as the degree to which all detected nuclei are fully enclosed by cytoplasm. This reward penalizes the failure mode of generative models wherein channel-level structures become spatially inconsistent:

\begin{equation}
\label{eq:nucincyto}
r_\text{Nuc-in-Cyto}(\hat{x}_1, c) = \frac{\text{area}(\text{nucleus mask} \cap \text{cytoplasm mask})}{\text{area}(\text{cytoplasm mask})}
\end{equation}

\paragraph{Nuclear roundness.}
Nuclear shape is a biologically informative feature that varies systematically across MoA classes. For instance, microtubule destabilizers produce fragmented, irregular nuclei, while other perturbations preserve smooth, round nuclear morphology. We compute the roundness reward as:
\begin{equation}
\label{eq:roundness}
r_\text{Roundness}(\hat{x}_1, c) = -\Bigl[\frac{1}{N_{\text{Nu}}}\sum_{i=1}^{N_{\text{Nu}}}\frac{4\pi \cdot \text{area}_{i}}{\text{perimeter}_{i}^2} - \mu^{(y_c)}\Bigr]^2 / [\sigma^{(y_c)}]^2,
\end{equation}
where $N_{\text{Nu}}$ is the number of identified nuclei, and $\mu^{(y_c)}$ and $\sigma^{(y_c)}$ are the average and standard deviation roundness of the MoA-conditioned ground truth distribution of $\hat{x}_1$ with label $y_c$.
This encourages the model to produce nuclear shapes consistent with the known morphological effects of each 
perturbation category.

\subsubsection{Morphological Statistics.}
\label{sec:reward_morphological}

These rewards ensure that the size and density of cellular components in generated images match those of the ground truth distribution, conditioned on the MoA. Unlike the structural rewards above, which enforce qualitative relationships, these rewards target quantitative statistics of the cell population.

For each of the following statistics $s$---maximum nucleus size, maximum cytoplasm size, nucleus count, and cytoplasm count---we compute the value $s(\hat{x}_1)$ from the generated image and compare it to the ground truth distribution for the corresponding MoA class. Let $\mu_s^{(y_c)}$ and $\sigma_s^{(y_c)}$ denote the mean and standard deviation of statistic $s$ across real images with MoA label $y_c$. We define 
the reward as the negative normalised deviation:
\begin{equation}
\label{eq:reward_morphological}
r_s(\hat{x}_1, c) = -\frac{\bigl[s(\hat{x}_1) - \mu_s^{(y_c)}\bigr]^2}{\bigl[\sigma_s^{(y_c)}\bigr]^2}, s\in\{\text{NucSize}, \text{CytoSize}, \text{NucCount}, \text{CytoCount}\}.
\end{equation}
This formulation penalises generated images whose statistics deviate from the ground truth mean, scaled by 
the natural variability within each MoA class. The four statistics capture complementary aspects of cell 
morphology:

\begin{itemize}
    \item \textbf{Maximum nucleus size} and \textbf{maximum cytoplasm size} reflect the largest cellular 
    components in the image, capturing whether the model produces cells of appropriate scale for the given 
    perturbation. We use the maximum rather than the average to mitigate the potential issue of partially visible cells biasing the statistics.
    \item \textbf{Nucleus count} and \textbf{cytoplasm count} measure cell density, ensuring the model does 
    not hallucinate or omit cells relative to what is observed under each treatment condition.
\end{itemize}

\noindent Together, these four rewards encourage the generated cell population to be statistically 
consistent with real observations in terms of both individual cell size and overall cell density.

\begin{figure}[!tb]
    \centering
    \includegraphics[width=0.95\linewidth]{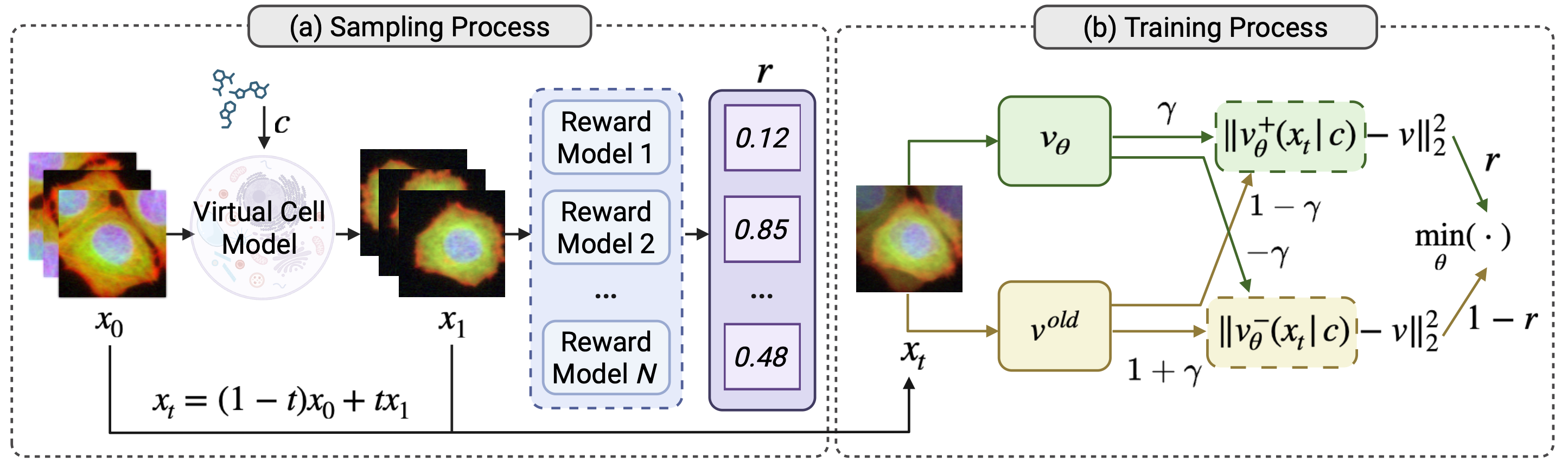}
    \caption{\textbf{\emph{CellFluxRL} algorithm.} RL post-training seeks to increase the likelihood of high-reward samples and decrease the likelihood of low-reward samples. Therefore, the core training loop of \emph{CellFluxRL} consists of interleaved phases of sampling and training. (a) Sampling: we generate multiple rollouts from a fixed control image and perturbation condition, scoring each with the reward models. (b) Training: because exact likelihoods in flow matching are intractable, we construct positive and negative velocities from the batch of rollouts and optimize them contrastively to achieve this goal, following DiffusionNFT~\cite{zheng2025diffusionnft}.}
    \label{fig:method}
\end{figure}

\subsection{Reinforcement Learning on Biological Rewards}
\label{sec:rl_algorithm}

We now describe how we optimize the pretrained base model $v_\theta$ with respect to the reward functions 
introduced in \S\ref{sec:reward_models}.

\paragraph{Algorithm overview.}
Our objective is to maximize the combined reward defined in Eq.\eqref{eq:rl_objective} (\S\ref{sec:reward_driven_generation}). Since these biological reward functions are non-differentiable, standard backpropagation is inapplicable, necessitating an RL approach. The core principle of our RL strategy is to increase the generation likelihood of high-reward samples while penalizing low-reward ones. 
We adopt DiffusionNFT~\cite{zheng2025diffusionnft}, a state-of-the-art online RL algorithm for flow matching. It operates on the flow's forward process, avoiding intractable log likelihoods, and is constructed from \textit{distribution-agnostic} components, hence extending naturally to our source-to-target flow matching setting without modification.


\paragraph{Algorithm detail.}
At each iteration, DiffusionNFT collects a batch of generated images, evaluates them with respect to the reward functions, and uses the rewards to define an improvement direction over the current policy. The key idea is to split generated samples into \textit{positive} (\emph{high-reward}) and \textit{negative} (\emph{low-reward}) subsets and learn a contrastive update that moves the model towards the positive distribution. Concretely, given a generated image $\hat{x}_1$ with optimality reward $r \in [0,1]$, 
the training objective is:
\begin{equation}
\label{eq:nft_loss}
\begin{split}
\mathcal{L}(\theta) = &\mathbb{E}_{c,\, \pi^{\mathrm{old}}(\hat{x}_1 | c, x_0),\, t} \Big[ r \, \| v_\theta^+(x_t, c, t) - v \|_2^2 + \; (1 - r) \, \| v_\theta^-(x_t, c, t) - v \|_2^2 \Big]\\
&+ \beta \, D_\text{KL}\!\left( v_{\theta} \,\|\, v^{\mathrm{old}} \right),
\end{split}
\end{equation}
where $\beta$ is the KL divergence weight, $x_t = \alpha_t x_0 + \gamma_t x_1$ is the forward-noised version of the generated image, 
$v = \dot{\alpha}_t x_0 + \dot{\gamma_t} x_1$ is the corresponding velocity target, and 
$v_\theta^+$, $v_\theta^-$ are \textit{implicit} positive and negative policies defined as:
\begin{align}
v_\theta^+(x_t, c, t) &:= (1 - \gamma)\, v^{\mathrm{old}}(x_t, c, t) + \gamma\, v_\theta(x_t, c, t) \label{eq:implicit_pos} \\
v_\theta^-(x_t, c, t) &:= (1 + \gamma)\, v^{\mathrm{old}}(x_t, c, t) - \gamma\, v_\theta(x_t, c, t). \label{eq:implicit_neg}
\end{align}
Here $v^{\mathrm{old}}$ is the data-collection policy (a lagging copy of $v_\theta$), and $\gamma > 0$ is a hyperparameter controlling the guidance strength. The implicit parameterization is central to the algorithm: rather than training separate positive and negative models, a single policy $v_\theta$ is optimized such that its mixture with $v^{\mathrm{old}}$ simultaneously fits high-reward samples (via $v_\theta^+$) and avoids low-reward ones (via $v_\theta^-$). The optimal solution satisfies $v_{\theta^*} = v^{\mathrm{old}} + \frac{2}{\gamma} \Delta$, where $\Delta$ is the reinforcement guidance direction pointing from the negative towards the positive distribution. 

This formulation naturally regularizes the post-trained model towards the pretrained policy: when $\gamma$ is large, the guidance strength $\frac{2}{\gamma}$ is small and the model stays close to $v^{\mathrm{old}}$; when $\gamma$ is small, the model is allowed to deviate more aggressively. The data-collection policy $v^{\mathrm{old}}$ is itself updated via an exponential moving average of 
$v_\theta$.

\paragraph{Rollout and advantage estimation.}
During sampling, we fix a perturbation condition $c$ and a source control image $x_0$, and generate a group of $m$ candidate images $\{\hat{x}_1^{(i)}\}_{i=1}^m$. Since the base model, $v_\theta$, augments $x_0$ with Gaussian noise before transporting it via the learned velocity field, diversity within each group arises primarily from this stochastic noise injection as well as any ODE discretization error. We find this yields sufficient variation for DiffusionNFT to distinguish positive from negative generations. Each candidate is scored by the reward functions, and the raw rewards are normalized within the group to obtain optimality probabilities $r^{(i)} \in [0, 1]$, following the advantage normalization scheme~\cite{zheng2025diffusionnft}. The forward process is then applied to each generated image, and 
the loss in Eq.\eqref{eq:nft_loss} is computed over the group.

\subsection{Test-Time Scaling}
\label{sec:test_time_scaling}

A key benefit of having explicit reward functions is that they can not only be used for training, but also to select among candidate generations at inference time. This strategy has proven highly effective in reasoning models, where best-of-$N$ selection with a reward model or verifier yields consistent improvements that scale predictably with $N$~\cite{cobbe2021training, lightman2023let, snell2024scaling}. We apply the same principle to virtual cell generation.

Given a perturbation condition $c$ and a source control image $x_0$, we generate $N$ candidate images $\{\hat{x}_1^{(i)}\}_{i=1}^N$ and select the one with the highest reward:
\begin{equation}
\label{eq:best_of_n}
\hat{x}_1^* = \arg\max_{i \in \{1, \dots, N\}} r(\hat{x}_1^{(i)}, c)
\end{equation}
This provides a simple, training-free mechanism to improve generation quality given additional inference compute budget. Moreover, best-of-$N$ selection is complementary to reinforcement learning: 
RL post-training improves the \textit{base distribution} from which candidates are drawn, so that even modest 
values of $N$ yield high-quality outputs, while test-time selection contributes additional gains.


%% file: secs/4_exp_arxiv.tex
\section{Results}


In this section, we evaluate \emph{CellFluxRL}'s ability to generate biologically faithful cellular images. After detailing our experimental setup (\S\ref{sec:exp_details}), we present the main results, demonstrating that RL with biologically-aligned rewards systematically improves performance across all biological metrics (\S\ref{sec:results_main}). We show that test-time scaling enhances performance (\S\ref{sec:results_tts}) and provide ablation studies.

\subsection{Experimental details}
\label{sec:exp_details}

\paragraph{Datasets.} We follow the dataset configuration of the base \emph{CellFlux} model and use the high-content microscopy perturbation dataset BBBC021~\cite{caie2010high}. BBBC021 is a chemical perturbation dataset comprising 98K three-channel images at 96$\times$96 resolution, collected under 26 chemical perturbations grouped into 12 modes of action (MoA). 
We follow the train/test split used in CellFlux~\cite{zhang2025cellflux}.

\paragraph{Baselines.} We compare our proposed method against the pretrained base model, \emph{CellFlux}~\cite{zhang2025cellflux}, as well as two prior generative baselines, \emph{PhenDiff}~\cite{bourou2024phendiff} and \emph{IMPA}~\cite{palma2023predicting}.

\paragraph{Evaluation metrics.} In addition to the rewards reported in \S\ref{sec:reward_models} (higher is better), which measure the biological meaningfulness of the generated cells, we also report standard image quality metrics (FID and KID). These metrics measure the image distribution similarity via Frechet and kernel-based distances between generated images and ground truth images (lower is better).

\begin{table*}[!tb]
\caption{Quantitative comparison across biological rewards and generative quality metrics. Each row 
    represents a different evaluation metric, and each column represents a different method. TTS is \emph{CellFluxRL} with test time scaling by best-of-$N$ with $N=4$, where the best sample is selected by the weighted total reward. \textbf{Bold} values indicate the best performance; \underline{underlined} values indicate the second best.}
\rowcolors{2}{white}{light-light-gray}
\scriptsize
\centering
\setlength\tabcolsep{8.8pt}
\vspace{6pt}
\renewcommand{\arraystretch}{1.2}
    \begin{tabular}{lR{1.4cm}R{1.4cm}R{1.4cm}R{1.4cm}R{1.4cm}}
        \toprule
        \textbf{Metric} & \textbf{PhenDiff}~\cite{bourou2024phendiff} & \textbf{IMPA}~\cite{palma2023predicting} & \textbf{CellFlux}~\cite{zhang2025cellflux} & \textbf{CellFluxRL} & \textbf{\textit{+TTS}} \\
        \midrule
        \multicolumn{6}{c}{\textit{Biological Rewards}} \\
        MoA & 0.18 & 0.12 & 0.26 & \underline{0.34} & \textbf{0.56} \\
        Nuc-in-Cyto & 0.91 & 0.79 & 0.88 & \underline{0.96} & \textbf{0.97} \\
        Roundness & -0.24 & -0.32 & -0.34 & \underline{-0.26} & \textbf{-0.19} \\
        NucSize & \underline{-0.88} & -1.02 & -2.21 & -1.04 & \textbf{-0.38} \\
        CytoSize & -0.97 & -1.59 & -1.09 & \underline{-0.65} & \textbf{-0.41} \\
        NucCount & -2.61 & -1.05 & -0.83 & \underline{-0.53} & \textbf{-0.28} \\
        CytoCount & -3.22 & -1.39 & -1.03 & \underline{-0.68} & \textbf{-0.33} \\
        Overall & -5.20 & -3.19 & -2.44 & \underline{0.46} & \textbf{3.15} \\
        \midrule
        \multicolumn{6}{c}{\textit{Generation Quality}} \\
        FID & 41.94 & 35.70 & \textbf{20.36} & 24.01 & \underline{23.19} \\
        KID & 0.028 & 0.029 & \underline{0.015} & \textbf{0.014} & 0.016 \\
        \bottomrule
    \end{tabular}
    \label{tab:main_rewards}
\end{table*}

\paragraph{Training details.} We instantiate our method using a pretrained \emph{CellFlux} flow matching 
model. \emph{CellFluxRL} is post-trained using the online RL algorithm 
DiffusionNFT~\cite{zheng2025diffusionnft} to optimize a weighted sum of 7 biologically-constrained 
rewards: $r = 5.0\,r_\text{MoA} + 2.0\,r_\text{Nuc-in-Cyto} + r_\text{Roundness} + r_\text{NucSize} + r_\text{CytoSize} + r_\text{NucCount} + r_\text{CytoCount}$. 
MoA and Nuc-in-Cyto are empirically harder to optimize, motivating their higher weights.
The training objective includes a KL divergence weight ($\beta=1$) to regularize the post-trained
model against the pretrained policy and prevent reward hacking. All other hyperparameter settings follow DiffusionNFT~\cite{zheng2025diffusionnft}. At inference, we further 
enhance performance using test-time scaling (TTS) with a best-of-$N$ selection strategy 
($N = 4$), selecting the generated sample with the highest overall reward.  Training is 
conducted for $1200$ steps on 1 H100 GPU for 32 hours.

\begin{figure}[!tb]
    \centering
    \includegraphics[width=0.98\linewidth]{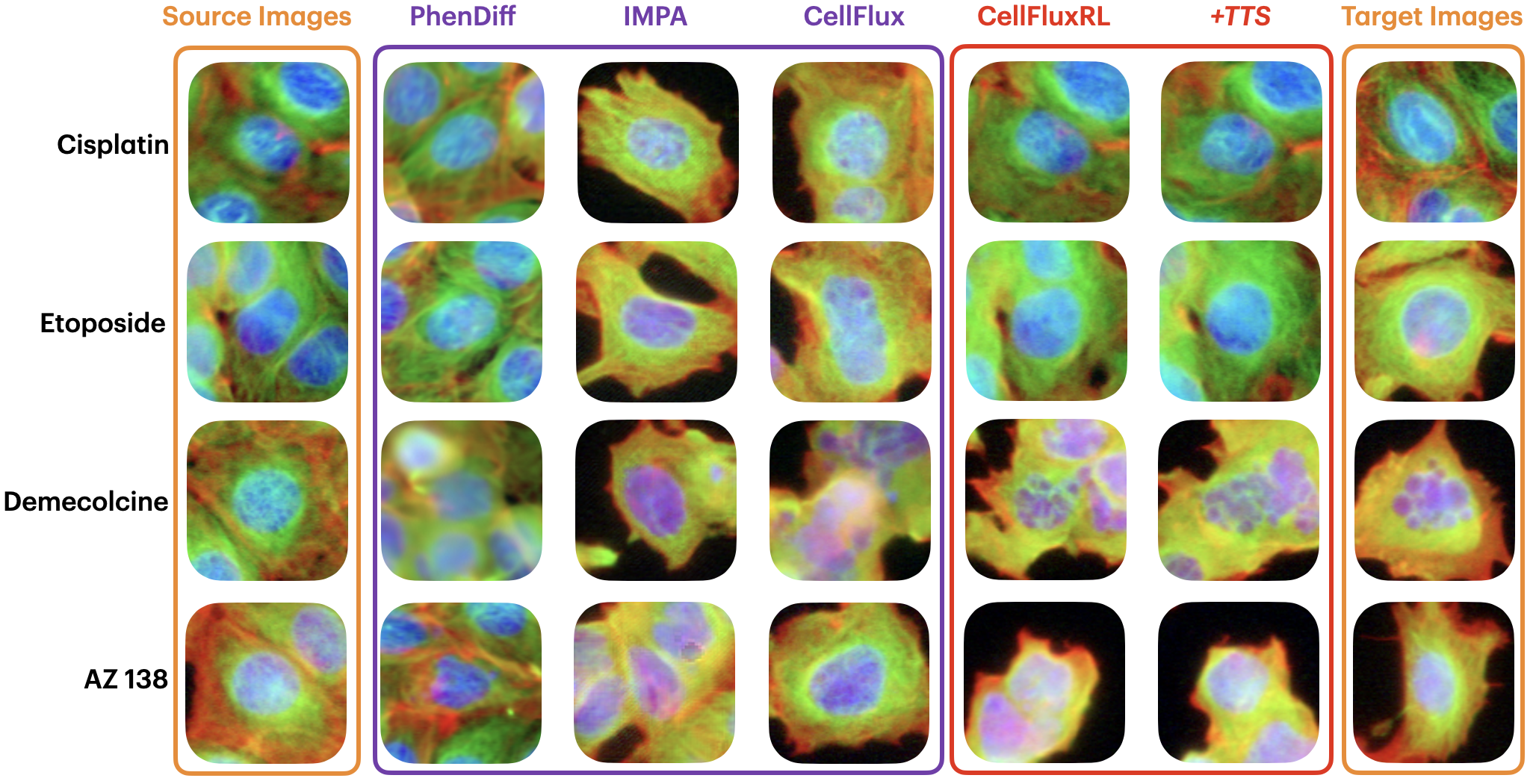}
    \caption{\textbf{Qualitative comparisons.} \emph{CellFluxRL} generates more biologically-grounded images, better capturing drug-induced morphological changes. In these examples, Etoposide-induced cell rounding, Demecolcine-driven microtubule destabilization, and AZ138-associated cell shrinkage are all more faithfully reproduced, and cell density more closely matches the ground truth for Cisplatin. Test-time scaling (+TTS) further refines these predictions toward the target images.}
    \label{fig:qual}
\end{figure}

\subsection{Reinforcement Learning Leads to Better Results}
\label{sec:results_main}

We compare \emph{CellFluxRL} against baselines on a suite of biological, structural, and morphological metrics, with results summarized in Table~\ref{tab:main_rewards}.
\emph{CellFluxRL} outperforms its pretrained counterpart \emph{CellFlux} across all reward metrics, confirming that RL post-training successfully steers the generative model toward greater physical correctness. The overall reward improves from $-4.44$ to $-1.54$, with gains spanning all three reward categories: biological function, structural constraints, and morphological statistics. 

\emph{CellFluxRL} rewards also surpass those of \emph{PhenDiff} and \emph{IMPA}. The improvements are particularly pronounced on MoA, where \emph{CellFluxRL}'s reward of $0.34$ substantially exceeds both \emph{PhenDiff} ($0.18$) and \emph{IMPA} ($0.12$). To give interpretable context to these numbers, the corresponding MoA classification accuracies---the fraction of generated images whose predicted mode of action matches the ground truth label---are $0.56$, $0.53$, $0.61$, and $0.66$ for \emph{PhenDiff}, \emph{IMPA}, \emph{CellFlux}, and \emph{CellFluxRL} respectively, indicating that \emph{CellFluxRL} produces images whose morphological profiles more faithfully distinguish different drug classes. Applying best-of-$N$ selection with $N=4$ to \emph{CellFluxRL} (+TTS) yields further improvements across all metrics, pushing the overall reward from $-1.54$ to $1.17$. 
We analyze test-time scaling behavior in further detail in \S\ref{sec:results_tts}.

In terms of generative quality, while not a primary optimization objective, \emph{CellFluxRL} achieves a state-of-the-art KID score and a competitive FID score compared to baselines.
The quantitative gains are corroborated qualitatively in Figure~\ref{fig:qual}. The baselines generate images that fail to reflect the expected 
biological response to each perturbation. \emph{CellFluxRL} consistently corrects these failures, 
and test-time scaling pushes generations further toward the ground truth.

\subsection{Test-time Scaling Further Enhances the Results}
\label{sec:results_tts}

\begin{figure}[!tb]
    \centering
    \includegraphics[width=\linewidth]{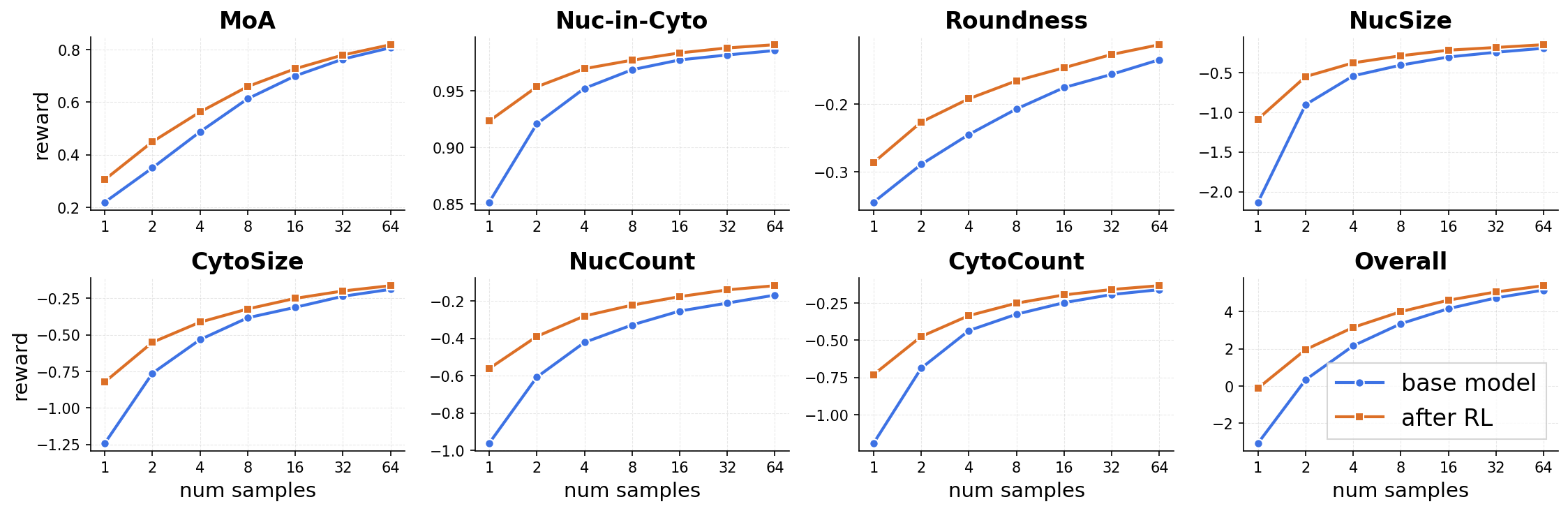}
    \vspace{-1em}
    \caption{Test-time scaling by best-of-$N$ further improves generation quality. The sample achieving the highest overall (combined) reward is selected from $N$ rollouts, and each individual reward is plotted. RL (orange) consistently exhibits better scaling than the base model (blue) across all rewards.}
    \label{fig:test-time}
\end{figure}

Figure~\ref{fig:test-time} shows how reward scores scale with the number of candidate samples $N$ for both the base model (\emph{CellFlux}) and the RL-post-trained model (\emph{CellFluxRL}). We generate $N$ candidates and select the one with the highest overall reward, the same weighted combination of individual rewards used during RL post-training, then report the individual reward metrics of the selected samples.
Both models exhibit monotonic improvements as $N$ increases across all metrics.
RL post-training improves the scaling efficiency: the \emph{CellFluxRL} curve sits consistently above the base model curve across all metrics, meaning that for any fixed compute budget (i.e., any given $N$), the RL-post-trained model achieves higher reward. Because each sample drawn from the improved distribution is more likely to be physically correct, selection is more efficient. The effect is especially pronounced at small $N$, where the quality of the base distribution matters most.


\subsection{Sensitivity Analysis on Kullback–Leibler Divergence Weight}

\begin{figure}[!tb]
    \centering
    \includegraphics[width=\linewidth]{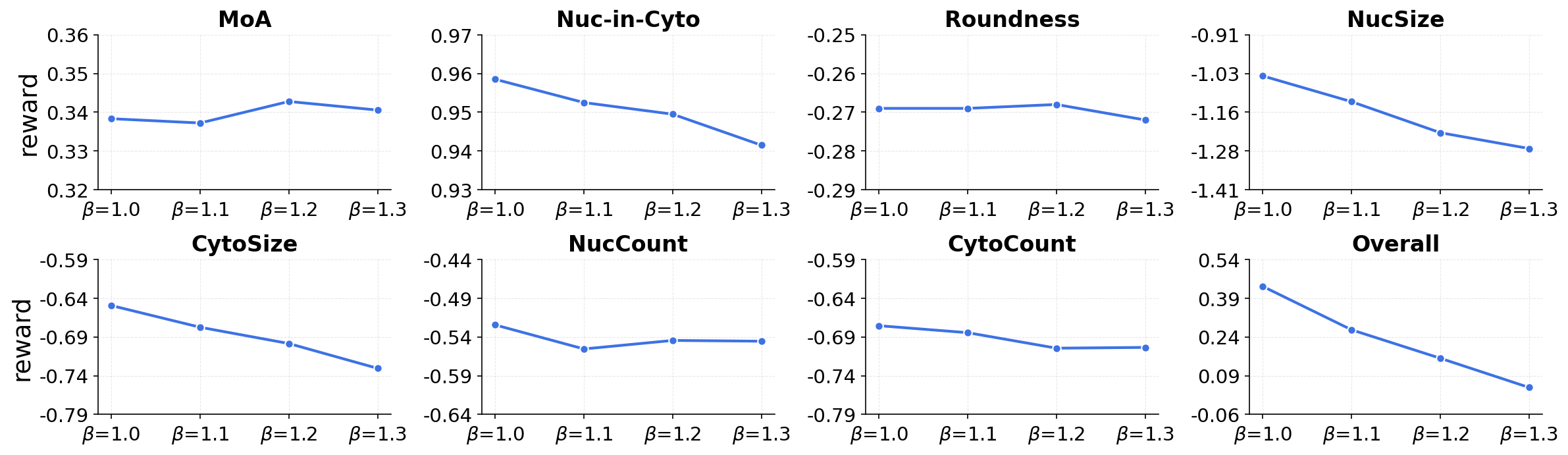}
    \caption{Sensitivity analysis on KL weight $\beta$. Each subplot shows reward sensitivity to $\beta$ after RL post-training.}
    \label{fig:study_kl}
\end{figure}

We study the sensitivity of final model performance on the KL divergence weight, $\beta$, varying it from 1.0 
to 1.3. The results, presented in 
Figure~\ref{fig:study_kl}, show a trade-off: structural and morphological 
correctness metrics favor a smaller KL weight, whereas biological function 
rewards slightly improve with a larger weight. This suggests that achieving 
higher biological accuracy requires more substantial model adjustments than 
improving structural or morphological fidelity.

\begin{table}[!tb]
\caption{Single-reward optimization with reinforcement learning. Each column shows a model optimized for one reward only or \emph{CellFluxRL} (all rewards), 
    evaluated across the metrics in each row. \textbf{Bold} values indicate the best performance; \underline{underlined} values indicate the second best; \dashuline{dashed-underlined} values indicate the third best.}
\rowcolors{2}{white}{light-light-gray}
\setlength\tabcolsep{1.7pt}
\vspace{6pt}
\renewcommand{\arraystretch}{1.2}
\scriptsize
\centering
    \begin{tabular}{lcccccccc}
        \toprule
        Metric & MoA & Nuc-in-Cyto & Roundness & NucSize & CytoSize & NucCount & CytoCount & \emph{CellFluxRL}\\
        \midrule
        MoA & \textbf{0.412} & 0.204 & 0.263 & \dashuline{0.301} & 0.212 & 0.279 & 0.277 & \underline{0.337} \\
        Nuc-in-Cyto & 0.852 & \textbf{0.977} & 0.907 & 0.904 & \underline{0.976} & 0.867 & 0.937 & \dashuline{0.956} \\
        Roundness & -0.342 & -0.332 & \textbf{-0.198} & \dashuline{-0.298} & -0.340 & -0.271 & -0.304 & \underline{-0.264} \\
        NucSize & -3.319 & \dashuline{-1.724} & -2.328 & \textbf{-0.702} & -1.784 & -2.909 & -2.853 & \underline{-1.038} \\
        CytoSize & -1.198 & \underline{-0.545} & -0.919 & -0.917 & \textbf{-0.497} & -1.113 & -0.801 & \dashuline{-0.652}\\
        NucCount & -0.760 & -0.904 & -0.685 & -0.818 & -0.881 & \textbf{-0.476} & \underline{-0.511} & \dashuline{-0.528} \\
        CytoCount & -0.922 & -1.074 & -0.905 & -0.946 & -1.063 & \underline{-0.629} & \textbf{-0.580} & \dashuline{-0.677} \\ 
        \bottomrule
    \end{tabular}
    \label{tab:study_single_reward}

\end{table}

\subsection{Ablation Optimizing for a Single Reward}

To isolate the effect of individual rewards, we conduct ablations optimizing for one reward at a time.
As shown in Table~\ref{tab:study_single_reward}, this targeted optimization 
significantly improves performance on the chosen metric, but the improvements 
on other metrics are limited. In contrast, \emph{CellFluxRL}, 
optimized with the combined weighted sum of all seven rewards, achieves either 
the second- or third-best score on every individual metric. This demonstrates that the combined reward strategy 
enables balanced, multi-faceted improvement across all biological, structural, 
and morphological criteria simultaneously.

\section{Conclusion}

In this work, we tackle a central challenge for virtual cells: encouraging generated images to be physically correct and biologically plausible. 
We accomplish this by introducing a suite of biologically-informed rewards, which serve a unified role---as evaluation metrics, as training signals for reinforcement learning, and as a tool for test-time scaling. 
We use these rewards to optimize the state-of-the-art perturbation model \emph{CellFlux}, yielding \emph{CellFluxRL}. 
Across all rewards, our approach consistently improves physical plausibility and maintains overall image quality over the base model. 
Together, these results evolve virtual cell generation from pixel-level realism toward biological groundedness.

\paragraph{Limitations and Future Work.} Our biological rewards are manually engineered based on domain expertise. A promising future direction is leveraging Large Language Models to automatically translate scientific literature into executable reward functions, significantly lowering the barrier to entry for novel domains. 
In this work, our core contribution is the modular multi-reward RL pipeline itself. As new generative architectures and more efficient RL algorithms emerge, they can be seamlessly integrated into this framework to further advance physically grounded generation.
Finally, while we demonstrate our RL pipeline on virtual cell modeling, our approach provides a domain-agnostic framework for constrained generation. It can be adapted to other scientific fields governed by strict, non-differentiable rules, such as material science or medical imaging. 

%% file: secs/X_appendix_arxiv.tex
\clearpage

\section{Algorithm of CellFluxRL}
\label{app:algorithm_cellfluxrl}

We present the full training procedure of \emph{CellFluxRL} in Algorithm~\ref{alg:cellfluxrl}. The algorithm adapts DiffusionNFT~\cite{zheng2025diffusionnft} to the source-to-target flow matching setting and replaces the generic reward with our suite of biologically grounded reward functions.

\noindent\textbf{Relation to DiffusionNFT.}
CellFluxRL adapts DiffusionNFT in three key ways.
(i)~\emph{Source-to-target flow}: unlike standard diffusion where the forward process adds Gaussian noise from a fixed prior, our forward process interpolates between the source control image $x_0$ and the generated target $\hat{x}_1$ (Line~11), matching the CellFlux flow matching formulation.
(ii)~\emph{Biological reward suite}: the scalar reward $r$ (Lines~6--8) is the weighted combination of $K$ biologically grounded evaluators covering MoA consistency, structural plausibility, and morphological statistics, rather than a single generic reward.
(iii)~\emph{Group rollouts conditioned on $(x_0, c)$ pairs}: each rollout group is conditioned on a fixed control image alongside a perturbation condition, so within-group diversity arises from the stochastic noise injection applied to $x_0$ by CellFlux, rather than from random latent initialization.

\begin{algorithm}[thp!]
\caption{\emph{CellFluxRL}: Biologically-Constrained RL Post-Training for Flow Matching}
\label{alg:cellfluxrl}
\begin{algorithmic}[1]
\Require Pretrained velocity $v_\theta^{\mathrm{ref}}$;\; biological rewards $\{r_k\}_{k=1}^{K}$ with weights $\{w_k\}$;\;
         dataset $\mathcal{D}_{\mathrm{data}} = \{(x_0, c)\}$;\; group size $m$;\;
         hyperparameters $\beta$ and $\beta_{\mathrm{KL}}$;\; learning rate $\lambda$
\State \textbf{Initialize:}\ data collection policy $v^{\mathrm{old}} \leftarrow v_\theta^{\mathrm{ref}}$;\quad training policy $v_\theta \leftarrow v_\theta^{\mathrm{ref}}$;\quad data buffer $\mathcal{D} \leftarrow \emptyset$
\For{each iteration $i = 1, 2, \ldots$}
    \Statex \hspace{1.2em}\textit{\# --- Phase 1: Rollout (Data Collection) ---}
    \For{each $(x_0, c)$ sampled from $\mathcal{D}_{\mathrm{data}}$}
        \State Generate $m$ candidate perturbed images $x_1$ by from the policy $\pi^{\mathrm{old}}$ (by integrating $v^{\mathrm{old}}(\cdot | c)$ from $x_0$):
               \[
                 \bigl\{\hat{x}_1^{(j)}\bigr\}_{j=1}^{m} \;\sim\; \pi^{\mathrm{old}}(\cdot \mid x_0,\, c)
               \]
        \State Compute combined biological reward for each candidate:
               \[
                 r^{(j)} \;=\; \textstyle\sum_{k=1}^{K} w_k\, r_k\!\bigl(\hat{x}_1^{(j)},\, c\bigr).
               \]
        \State Normalize within the group and map to optimality probability $r^{(j)} \in [0,1]$:
               \[
                 r^{(j)}_{\mathrm{norm}} = r^{(j)} - \tfrac{1}{m}\textstyle\sum_j r^{(j)},
                 \qquad
                 r^{(j)}_{\mathrm{adv}} = 0.5 + 0.5 \cdot \mathrm{clip}\!\Bigl(\tfrac{r^{(j)}_{\mathrm{norm}}}{Z_c},\,-1,\,1\Bigr),
               \]
               \Statex \hspace{2.9em} where $Z_c > 0$ is a normalization hyperparameter.
        \State Append $\bigl\{\bigl(x_0,\, c,\, \hat{x}_1^{(j)},\, r^{(j)}_{\mathrm{adv}}\bigr)\bigr\}_{j=1}^m$ to buffer $\mathcal{D}$.
    \EndFor
    \Statex \hspace{1.2em}\textit{\# --- Phase 2: Training (Gradient Step) ---}
    \For{each mini-batch $\bigl\{(x_0, c, \hat{x}_1, r_{\mathrm{adv}})\bigr\} \subset \mathcal{D}$}
        \State Sample timestep $t \sim \mathcal{U}[0,1]$
        \State Apply source-to-target forward process:\;
               $x_t = \alpha_t x_0 + \beta_t \hat{x}_1$,\quad
               $v_t = \dot{\alpha}_t x_0 + \dot{\beta}_t \hat{x}_1$
        \State Construct implicit positive and negative velocities:
               \begin{align*}
                 v_\theta^+\!(x_t,c,t) &= (1-\beta)\,v^{\mathrm{old}}(x_t,c,t) + \beta\,v_\theta(x_t,c,t) \\
                 v_\theta^-\!(x_t,c,t) &= (1+\beta)\,v^{\mathrm{old}}(x_t,c,t) - \beta\,v_\theta(x_t,c,t)
               \end{align*}
        \State Compute contrastive loss:
               \[
                 \mathcal{L}(\theta) =\;
                 r_{\mathrm{adv}}\bigl\|v_\theta^+ - v\bigr\|_2^2
                 + (1-r_{\mathrm{adv}})\,\bigl\|v_\theta^- - v\bigr\|_2^2
                 + \beta_{\mathrm{KL}} \, D_{\mathrm{KL}}\!\bigl(v_\theta \,\|\, v^{\mathrm{old}}\bigr)
               \]
        \State Update parameters: $\theta \;\leftarrow\; \theta - \lambda\,\nabla_\theta\,\mathcal{L}(\theta)$
    \EndFor
    \Statex \hspace{1.2em}\textit{\# --- Phase 3: Policy Update ---}
    \State EMA update of data collection policy:\; $\theta^{\mathrm{old}} \leftarrow \eta_i\,\theta^{\mathrm{old}} + (1-\eta_i)\,\theta$
    \State Clear buffer: $\mathcal{D} \leftarrow \emptyset$
\EndFor
\State \Return Post-trained velocity field $v_\theta$
\end{algorithmic}
\end{algorithm}

\section{Implementation Details}
\label{app:implementation_details}

\textbf{Training Configurations.} Our setup largely follows 
DiffusionNFT~\cite{zheng2025diffusionnft} with a learning 
rate of $2\times10^{-5}$. For each collected generated image, 
forward noising and loss computation are performed at the 
corresponding sampling timesteps. We employ a Heun (2nd-order) 
ODE sampler for data collection.

\section{Dataset Details}
\label{app:dataset_details}

\textbf{BBBC021 dataset.}  
We use the BBBC021v1 image set~\cite{ljosa2012annotated,caie2010high} from the 
Broad Bioimage Benchmark Collection, a standard benchmark for 
fluorescence-microscopy-based phenotypic profiling. The dataset captures 
chemical perturbations in MCF-7 human breast cancer cells: approximately 
97,500 three-channel fluorescence images stained for DNA, F-actin, and 
beta-tubulin, covering 113 small-molecule compounds administered at eight 
concentrations each. The compounds span a broad range of biological mechanisms, 
and each is annotated with a mode-of-action (MoA) label. Following prior 
work~\cite{zhang2025cellflux}, we preprocess images by correcting 
illumination, cropping to $96\times96$ patches centered on nuclei, and 
filtering out low-quality frames, yielding approximately 98K images across 
26 perturbation conditions. Table~\ref{tab:moa} lists the MoA assignment 
for every compound used in our experiments.

\begin{table}[p!]
    \centering
    \caption{Modes of action (MoA) for compounds in BBBC021.}
    \small
    \rowcolors{2}{white}{light-light-gray}
    \setlength\tabcolsep{6pt}
    \vspace{6pt}
    \renewcommand{\arraystretch}{1.1}
    \begin{tabular}{ll}
        \toprule
        \textbf{Compound} & \textbf{MoA} \\
        \midrule
        Cytochalasin B & Actin disruptors \\
        Cytochalasin D & Actin disruptors \\
        Latrunculin B & Actin disruptors \\
        AZ258 & Aurora kinase inhibitors \\
        AZ841 & Aurora kinase inhibitors \\
        Mevinolin/Lovastatin & Cholesterol-lowering \\
        Simvastatin & Cholesterol-lowering \\
        Chlorambucil & DNA damage \\
        Cisplatin & DNA damage \\
        Etoposide & DNA damage \\
        Mitomycin C & DNA damage \\
        Camptothecin & DNA replication \\
        Floxuridine & DNA replication \\
        Methotrexate & DNA replication \\
        Mitoxantrone & DNA replication \\
        AZ138 & Eg5 inhibitors \\
        PP-2 & Epithelial \\
        Alsterpaullone & Kinase inhibitors \\
        Bryostatin & Kinase inhibitors \\
        PD-169316 & Kinase inhibitors \\
        Colchicine & Microtubule destabilizers \\
        Demecolcine & Microtubule destabilizers \\
        Nocodazole & Microtubule destabilizers \\
        Vincristine & Microtubule destabilizers \\
        Docetaxel & Microtubule stabilizers \\
        Epothilone B & Microtubule stabilizers \\
        Taxol & Microtubule stabilizers \\
        ALLN & Protein degradation \\
        Lactacystin & Protein degradation \\
        MG-132 & Protein degradation \\
        Proteasome inhibitor I & Protein degradation \\
        Anisomycin & Protein synthesis \\
        Cyclohexamide & Protein synthesis \\
        Emetine & Protein synthesis \\
        DMSO & DMSO \\
        \bottomrule
    \end{tabular}
    \label{tab:moa}
\end{table}

\section{Qualitative Comparison}
\label{app:qualitative_comparison}

\begin{figure}[p!]
    \centering
    \includegraphics[width=\textwidth]{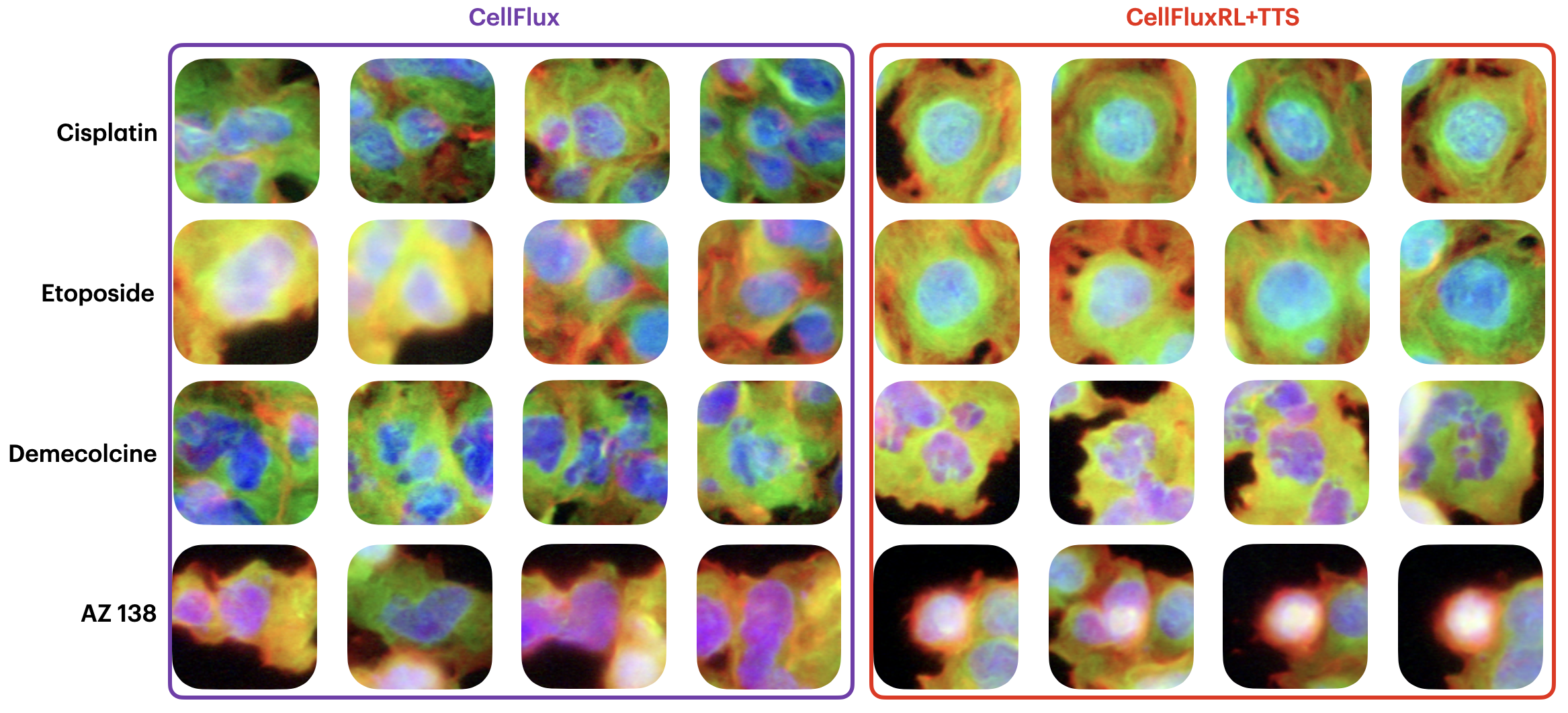}
    \caption{\textbf{MoA reward failure cases.} The pretrained CellFlux baseline (left) generates images that do not match the expected morphological profile for the given perturbation, as measured by the MoA reward. \emph{CellFluxRL + TTS} (right) corrects these failures by explicitly optimizing for MoA consistency during RL post-training. Ground-truth target images for the same perturbation conditions are shown in Figure~\ref{fig:qual} of the main paper.}
    \label{fig:qualitative_comparison_moa}
\end{figure}

\begin{figure}[p!]
    \centering
    \includegraphics[width=\textwidth]{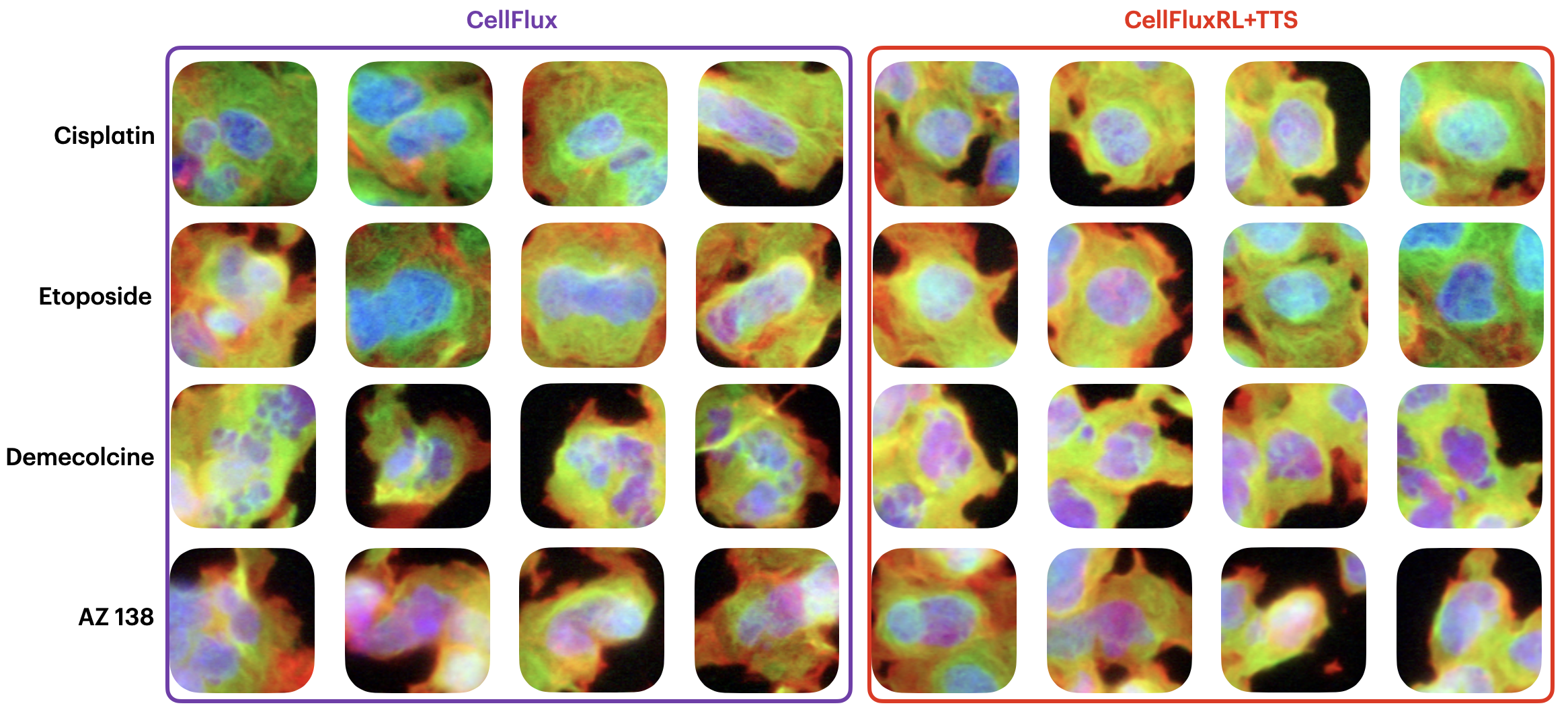}
    \caption{\textbf{Roundness reward failure cases.} The pretrained CellFlux baseline (left) produces nuclei with irregular, implausible shapes that deviate from the MoA-conditioned ground-truth distribution. \emph{CellFluxRL + TTS} (right) generates nuclei with roundness statistics consistent with real cells under the same perturbation condition. Ground-truth target images for the same perturbation conditions are shown in Figure~\ref{fig:qual} of the main paper.}
    \label{fig:qualitative_comparison_roundness}
\end{figure}

Figures~\ref{fig:qualitative_comparison_moa} 
and~\ref{fig:qualitative_comparison_roundness} provide 
additional qualitative comparisons between 
\emph{CellFluxRL + TTS} and the pretrained CellFlux 
baseline. Figure~\ref{fig:qualitative_comparison_moa} 
highlights cases where the base model fails to reproduce 
the perturbation-specific morphological profile expected 
for the given MoA---\emph{CellFluxRL + TTS} recovers 
the correct morphological signature by optimizing the 
MoA reward. Figure~\ref{fig:qualitative_comparison_roundness} 
shows cases where the base model produces nuclei 
with implausible shapes; by optimizing the roundness 
reward, \emph{CellFluxRL + TTS} generates nuclei whose 
shape is consistent with the ground-truth MoA-conditioned 
distribution.

\section{Normalized Reward Analysis}
\label{app:normalized_reward_analysis}

Table~\ref{tab:normalized_rewards} reports min-max normalized scores
across all 7 reward components.
\emph{CellFluxRL + TTS} consistently outperforms the pretrained CellFlux
baselines on every reward without requiring manual weight tuning.

\begin{table*}[t]
\caption{Quantitative comparison across biological evaluation metrics (rows) for CellFlux baselines and \emph{CellFluxRL} variants (columns), where \emph{CellFluxRL} is post-trained with min-max normalized rewards. \textbf{Bold} values indicate the best performance per metric.}
\rowcolors{2}{white}{light-light-gray}
\scriptsize
\centering
\setlength\tabcolsep{8.8pt}
\vspace{6pt}
\renewcommand{\arraystretch}{1.2}
    \begin{tabular}{lR{2.8cm}R{2.8cm}}
        \toprule
        \textbf{Metric}  & \textbf{CellFlux}~\cite{zhang2025cellflux} & \textbf{CellFluxRL}  \\
        \midrule
        \multicolumn{3}{c}{\textit{Biological Rewards}} \\
        MoA &  0.26  & \textbf{0.28} \\
        Nuc-in-Cyto &  0.88 & \textbf{0.96} \\
        Roundness &  -0.34 & \textbf{-0.25} \\
        NucSize &  -2.21 & \textbf{-1.34} \\
        CytoSize & -1.09 & \textbf{-0.64} \\
        NucCount & -0.83  & \textbf{-0.46} \\
        CytoCount & -1.03  & \textbf{-0.62} \\
        \bottomrule
    \end{tabular}
    \label{tab:normalized_rewards}
\end{table*}

\section{Held-out Reward Evaluation}
\label{app:heldout_reward_evaluation}

While using the training reward for evaluation is common practice in
RL post-training of diffusion models, reward hacking remains a potential
risk in any RL post-training pipeline. To assess whether the improvements
of \emph{CellFluxRL} reflect genuine biological gains rather than
exploitation of reward-implementation-specific idiosyncrasies, we
evaluate using two independent reward functions unseen during
RL training.

First, we trained two held-out MoA classifiers with architectures
(CLIP~\cite{radford2021learning}; DINOv2~\cite{oquab2023dinov2}) divergent from the paper's Inceptionv3 classifier.
As shown in Table~\ref{tab:heldout_rewards}, \emph{CellFluxRL}
consistently improves over CellFlux under both held-out classifiers
(CLIP: $0.45 \rightarrow 0.52$; DINOv2: $0.34 \rightarrow 0.40$),
corroborating the trend measured with the training classifier.
Second, to test whether the observed structural improvements depend on
Cellpose-specific segmentation results, we evaluate using an independent
segmentation pipeline, StarDist~\cite{schmidt2018cell}. The nucleus-in-cytoplasm score again
improves substantially ($0.88 \rightarrow 0.95$), comparable to the
improvement measured with Cellpose ($0.88 \rightarrow 0.96$). These
results suggest the gains transfer to independent evaluators rather
than exploiting reward-implementation-specific idiosyncrasies.

\begin{table}[t]
\caption{Evaluation with held-out reward functions unseen during RL training. MoA and NuInCy use the original training-time evaluators
(Inceptionv3 classifier and Cellpose segmentation, respectively).
MoA-CLIP and MoA-DINO use independently trained held-out MoA classifiers,
and NuInCy-StarDist uses an independent segmentation pipeline.
\textbf{Bold} values indicate the best performance per metric.}
\rowcolors{2}{white}{light-light-gray}
\scriptsize
\centering
\setlength\tabcolsep{6pt}
\renewcommand{\arraystretch}{1.2}
\resizebox{0.66\columnwidth}{!}{%
\begin{tabular}{lccccc}
\toprule
 & \textbf{MoA} & \textbf{MoA-CLIP} & \textbf{MoA-DINO} & \textbf{NuInCy} & \textbf{NuInCy-StarDist} \\
\midrule
\textbf{CellFlux}~\cite{zhang2025cellflux} & 0.26 & 0.45 & 0.34 & 0.88 & 0.88 \\
\textbf{CellFluxRL} & \textbf{0.34} & \textbf{0.52} & \textbf{0.40} & \textbf{0.96} & \textbf{0.95} \\
\bottomrule
\end{tabular}%
}
\label{tab:heldout_rewards}
\end{table}